\documentclass[10pt,twocolumn,letterpaper]{article}

\usepackage{cvpr}              

\usepackage{amsmath}
\usepackage{amssymb}
\usepackage{booktabs}

\usepackage{times,graphicx,amsmath,amssymb,booktabs,tabulary,multirow,overpic,xcolor,url}
\usepackage{colortbl}
\usepackage{makecell}
\usepackage{float}
\usepackage[accsupp]{axessibility}

\definecolor{sh_gray}{rgb}{0.84,0.84,0.84}
\definecolor{sh_gray2}{rgb}{1,0.89,0.75}
\definecolor{color3}{rgb}{0.95,0.95,0.95}
\definecolor{color4}{rgb}{0.96,0.96,0.86}
\definecolor{color5}{rgb}{0.90,0.90,0.90}

\usepackage[pagebackref,breaklinks,colorlinks]{hyperref}

\usepackage[capitalize]{cleveref}
\usepackage{multirow}
\crefname{section}{Sec.}{Secs.}
\Crefname{section}{Section}{Sections}
\Crefname{table}{Table}{Tables}
\crefname{table}{Tab.}{Tabs.}
\usepackage{algorithm}
\usepackage{algorithmic}

\def\cvprPaperID{70} 
\def\confName{CVPR NTIRE Workshop}
\def\confYear{2022}

\begin{document}

\title{MST++: Multi-stage Spectral-wise\\ Transformer for Efficient Spectral Reconstruction}

\author{Yuanhao Cai $^{1,*}$, Jing Lin $^{1,}$\thanks{Equal Contribution, $\dagger$ Corresponding Author} ~, Zudi Lin $^2$, Haoqian Wang $^{1,\dagger}$,\\  Yulun Zhang $^3$, Hanspeter Pfister $^2$, Radu Timofte $^{3,4}$,  Luc Van Gool $^{3}$ \\
	$^{1}$ Shenzhen International Graduate School, Tsinghua University, \\$^2$ Harvard University, $^3$ CVL, ETH Z\"{u}rich, $^4$ CAIDAS, JMU W\"urzburg
}
\maketitle

\begin{abstract}
Existing leading methods for spectral reconstruction (SR) focus on designing deeper or wider convolutional neural networks (CNNs) to learn the end-to-end mapping from the RGB image to its hyperspectral image (HSI). These CNN-based methods achieve impressive restoration performance while showing limitations in capturing the long-range dependencies and self-similarity prior. To cope with this problem, we propose a novel Transformer-based method, Multi-stage Spectral-wise Transformer (MST++),  for efficient spectral reconstruction. In particular, we employ Spectral-wise Multi-head Self-attention (S-MSA) that is based on the HSI spatially sparse while spectrally self-similar nature to compose the basic unit, Spectral-wise Attention Block (SAB). Then SABs build up Single-stage Spectral-wise Transformer (SST) that exploits a U-shaped structure to extract multi-resolution contextual information. Finally, our MST++, cascaded by several SSTs, progressively improves the reconstruction quality from coarse to fine. Comprehensive experiments show that our MST++ significantly outperforms other state-of-the-art methods. In the NTIRE 2022 Spectral Reconstruction Challenge,
our approach won the \textbf{First} place. Code and pre-trained models are publicly available at \url{https://github.com/caiyuanhao1998/MST-plus-plus}. 

\end{abstract}

\section{Introduction}

\begin{figure}[h]
	\begin{center}
		\begin{tabular}[t]{c} \hspace{-3.8mm} 
			\includegraphics[width=0.46\textwidth]{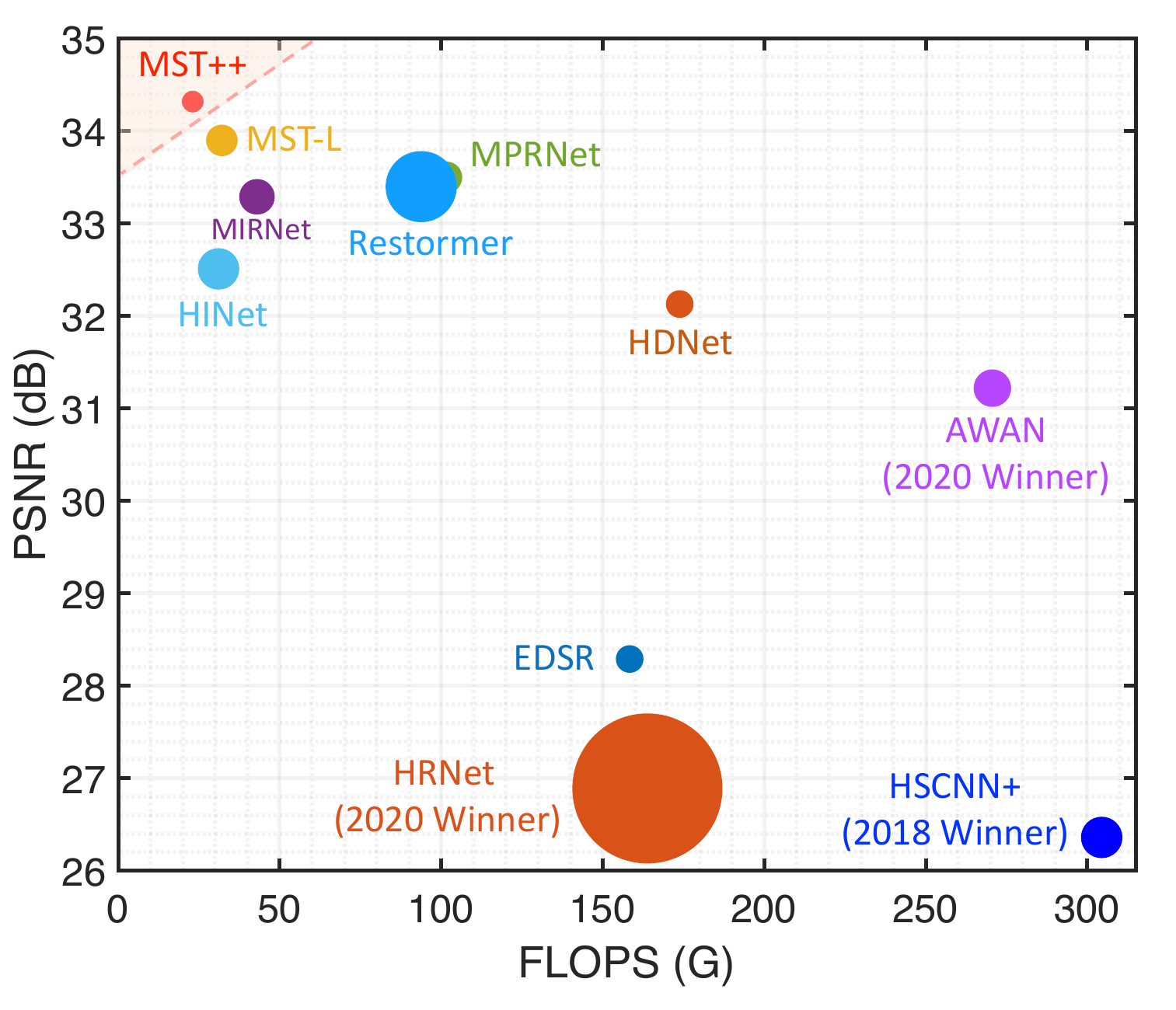}
		\end{tabular}
	\end{center}
	\vspace{-8mm}
	\caption{\small PSNR-Params-FLOPS comparisons with other spectral reconstruction algorithms. The horizontal axis is FLOPS (computational complexity), the vertical axis is PSNR ( performance), and the circle radius is Params (memory cost). Our Multi-stage Spectral-wise Transformer (MST++) surpasses other methods while requiring significantly cheaper FLOPS and Params. }
	\label{fig:teaser}
	\vspace{-5mm}
\end{figure}

Hyperspectral imaging records the real-world scene spectra in narrow bands, where each band captures the information at a specific spectral wavelength. Compared to normal RGB images, HSIs have more spectral bands to store richer information and delineate more details of the captured scenes. Because of this advantage, HSIs have wide applications such as medical image processing~\cite{mi_1,mi_2,mi_3}, remote sensing~\cite{rs_1,rs_2,rs_3}, object tracking~\cite{ot_1,ot_2}, and so on. Nonetheless, such HSIs with plentiful spectral information is time-consuming that spectrometers are used to scan the scenes along the
spatial or spectral dimension. This limitation impedes the application scope of HSIs, especially in dynamic or real-time scenes.

One way to solve this problem is to develop snapshot compressive imaging (SCI) systems and computational reconstruction algorithms~\cite{mst,pngan,hdnet,cst,gap_tv,gapnet,tsa_net,lambda,gsm,hssp,dnu,desci,twist,self,PDMSR} from 2D measurement to 3D HSI cube. Nevertheless, these methods rely on expensive hardware devices. To reduce costs, spectral reconstruction (SR) algorithms are proposed to reconstruct the HSI from a given RGB image, which can be easily obtained by RGB cameras. 

Conventional SR methods are mainly based on sparse coding or relatively shallow learning models. Nonetheless, these model-based methods suffer from limited representing capacity and poor generalization ability. Recently, with the development of deep learning, SR has witnessed significant progress. Deep convolutional neural networks (CNNs) have been applied to learn the end-to-end mapping function from  RGB images to HSI cubes. Although impressive performance have been achieved, these CNN-based methods show limitations in capturing long-range dependencies and inter-spectra self-similarity.

In recent years, the natural language processing (NLP) model, Transformer~\cite{vaswani2017attention}, has been applied in computer vision and achieved great success. The multi-head self-attention (MSA) mechanism in Transformer does better in modeling long-range dependencies and non-local self-similarity than CNN, which can alleviate the limitations of CNN-based SR algorithms. However, directly using standard Transformer~\cite{global_msa,liu2021swin} for SR will encounter two main issues. \textbf{(i)} Global~\cite{global_msa} and local~\cite{liu2021swin} Transformer captures inter-actions of spatial regions. Yet, the HSI representations are spatially sparse while spectrally highly self-similar. Thus,  modeling spatial inter-dependencies may-be less cost-effective than capturing inter-spectra correlations. \textbf{(ii)} On the one hand, the computational complexity of standard global MSA is quadratic to the spatial dimension, which is a huge burden that may be unaffordable. On the other hand, local window-based MSA suffers from limited receptive fields within position-specific windows. 

To address the aforementioned limitations, we propose the first Transformer-based framework, Multi-stage Spectral-wise Transformer (MST++) for efficient spectral reconstruction from RGB images. Notely, our MST++ is based on the prior work MST~\cite{mst}, which is customized for spectral compressive imaging restoration. \textbf{Firstly}, we note that HSI signals are spatially sparse while spectrally self-similar. Based on this nature, we adopt the Spectral-wise Multi-head Self-Attention (S-MSA) to compose the basic unit, Spectral-wise Attention Block (SAB). S-MSA  treats each spectral feature map as a token to calculate the self-attention along the spectral dimension. \textbf{Secondly}, our SABs build up our proposed Single-stage Spectral-wise Transformer (SST) that exploits a U-shaped structure to extract multi-resolution spectral contextural  information which is critical for HSI restoration. \textbf{Finally}, our MST++, cascaded by several SSTs, develops a multi-stage learning scheme to progressively improve the reconstruction quality from coarse to fine, which significantly boosts the performance.

The main contributions of this work are listed as follow.
\begin{itemize}
	\item We propose a novel framework, MST++, for SR. To the best of our knowledge, it is the first attempt to explore the potential of Transformer in this task. 
	\item We validate a series of natural image restoration models on this SR task. Toward them, we propose a Top-K multi-model ensemble strategy to improve the SR performance. Codes and pre-trained models of these methods are made publicly available to serve as a baseline and toolbox for further research in this topic.
	\item Quantitative and qualitative experiments demonstrate that our MST++ dramatically outperforms SOTA methods while requiring much cheaper Params and FLOPS. Surprisingly, our MST++ won the \textbf{First} place in NTIRE 2022 Spectral Reconstruction Challenge~\cite{arad2022ntirerecovery}.
\end{itemize}

\section{Related Work}

\subsection{Hyperspectral Image Aquisition}
Traditional imaging systems for collecting HSIs often adopt spectrometers to scan the scene along the spatial or spectral dimensions. Three main types of scanners including whiskbroom scanner, pushroom scanner, and band sequential scanner are often used to capture HSIs. These scanners have been widely used in detecting, remote sensing, medical imaging, and environmental monitoring for decades.  For example, pushbroom scanner and whiskbroom scanner have been used in satellite sensors~\cite{satellite_1,satellite_2} for photogrammetric and remote sensing. However, the scanning procedure usually requires a long time, which makes it unsuitable for measuring dynamic scenes. Besides, the imaging devices are usually too large physically to be plugged in portable platforms. To address these limitations, researchers have developed SCI systems~\cite{sci_1,sci_2,sci_3,sci_5,sci_6} to capture HSIs, where the 3D HSI cube is compressed into a single 2D measurement~\cite{Yuan_review}. Among these SCI systems, coded aperture snapshot spectral imaging (CASSI)~\cite{tsa_net,sci_2} stands out and forms one promising research direction. Nonetheless, the SCI systems remain prohibitively expensive to date for consumer grade use. Even "low-cost" SCI systems are often in the \$ 10K -  \$ 100K. Therefore, the SR topic has significant research and practical value. 

\begin{figure*}[t]
	\begin{center}
		\begin{tabular}[t]{c} \hspace{-3mm} 
			\includegraphics[width=1.0\textwidth]{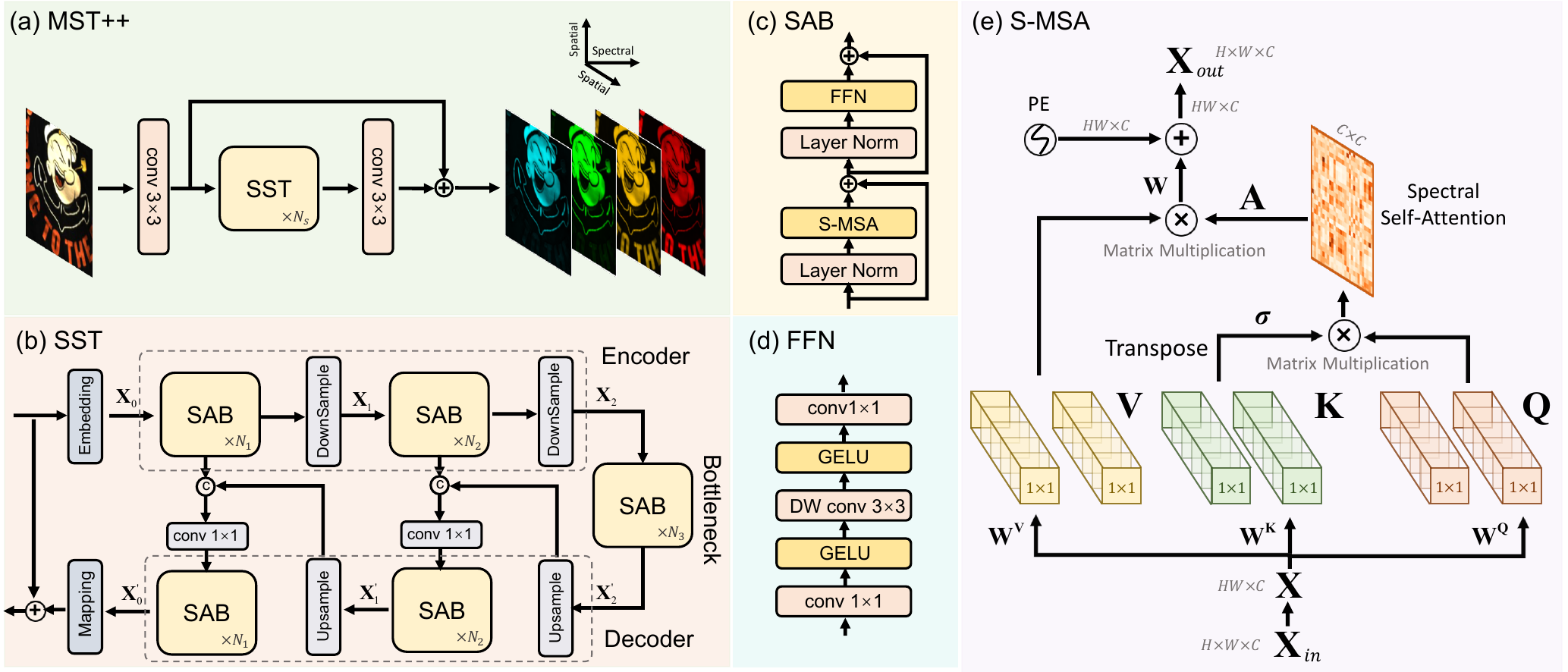}
		\end{tabular}
	\end{center}
	\caption{\small The overall pipeline of the proposed solution MST++. (a) Multi-stage Spectral-wise Transformer. (b) Single-stage Spectral-wise Transformer. (c) Spectral-wise Attention Block. (d) Feed Forward Network. (e) Spectral-wise Multi-head Self-Attention.}
	\label{fig:pipeline}
\end{figure*}

\subsection{Spectral Reconstruction from RGB}
Conventional SR methods~\cite{tradi_1,tradi_2,tradi_3,tradi_4,parmar2008spatio} are mainly based on hand-crafted hyperspectral priors. For instance,  Paramar \emph{et al.}~\cite{parmar2008spatio} propose a data sparsity expending method for HSI reconstruction. Arad \emph{et al.}~\cite{tradi_2} propose a sparse coding method that create a dictionary of HSI signals and their RGB projections. Aeschbacher \emph{et al.}~\cite{tradi_1} suggest using relatively shallow learning models from a specific spectral prior to fulfill spectral super-resolution. However, these model-based methods suffer from limited representing capacities and poor generalization ability. 

Recently, inspired by the great success of deep learning in natural image restoration~\cite{huang2020unfolding,rformer,fgst,pngan,li2022general,li2022dfan,li2021approaching,rdn,msfn,poan}, CNNs have been exploited to learn the underlying mapping function from RGB to HSI~\cite{xiong2017hscnn,shi2018hscnn,zhang2020pixel,stiebel2018reconstructing,galliani2017learned}. For instance, Xiong \emph{et al.}~\cite{xiong2017hscnn} propose a unified HSCNN framework for HSI reconstruction from both RGB images and compressive measurements. Shi \emph{et al.}~\cite{shi2018hscnn} use adapted residual blocks to build up a deep residual network HSCNN-R for SR. Zhang \emph{et al.}~\cite{zhang2020pixel} customize a pixel-aware deep function-mixture network consisting to model the RGB-to-HSI mapping. However, these CNN-based SR methods achieve impressive results but show limitations in capturing non-local self-similarity and long-range inter-dependencies.

\subsection{Vision Transformer}
\noindent The NLP model Transformer~\cite{vaswani2017attention} is  proposed for machine translation. In recent years, it has been introduced into computer vision and gained much popularity due to its advantage in capturing long-range correlations between spatial regions. In high-level vision, Transformer has been widely applied in image classification~\cite{liu2021swin,arnab2021vivit,global_msa,xcit,tc_2,tc_1}, object detection~\cite{de_detr,DETR,dy_detr,to_3,to_2,to_1}, semantic segmentation~\cite{tc_3,cao2021swin,SETR,ts_1,ts_2,ts_3}, human pose estimation~\cite{tokenpose,transpose,rsn,cai2019joint,udp++,prtr,th_1,th_2,th_3},  \emph{etc}. In addition, vision Transformer has also been used in low-level vision~\cite{mst,vsrt,ipt,rformer,swinir,fgst,cst}. For instance, Cai \emph{et al.}~\cite{mst} propose the first Transformer-based end-to-end framework MST for HSI reconstruction from compressive measurements. Lin \emph{et al.}~\cite{cst} embed the HSI sparsity into Transformer to establish a coarse-to-fine learning scheme for spectral comrpessive imaging. The prior work Uformer~\cite{uformer} adopts a U-shaped structure built up by Swin Transformer~\cite{liu2021swin} blocks for natural image restoration. Nonetheless, to the best of our knowledge, the potential of Transformer in spectral super-resolution has not been explored. This work aims to fill this research gap.

\section{Method}
\subsection{Network Architecture}
As shown in Fig.~\ref{fig:pipeline}, (a) depicts the proposed Multi-stage Spectral-wise Transformer (MST++), which is cascaded by $N_s$ Single-stage Spectral-wise Transformers (SSTs). Our MST++ takes a RGB image as the input and reconstructs its HSI counterpart. A long identity mapping is exploited to ease the training procedure. Fig.~\ref{fig:pipeline} (b) shows the U-shaped SST consisting of an encoder, a bottleneck, and a decoder. The embedding and mapping block are single $conv$3$\times$3 layers. The feature maps in the encoder sequentially undergo a downsampling operation (a strided $conv$4$\times$4 layer), $N_1$ Spectral-wise Attention Blocks (SABs), a downsampling operation, and $N_2$ SABs. The bottleneck is composed of $N_3$ SABs. The decoder employs a symmetrical architecture. The upsampling operation is a strided \emph{deconv}2$\times$2 layer. To avoid the information loss in the downsampling, skip connections are used between the encoder and decoder. Fig.~\ref{fig:pipeline} (c) illustrates the components of SAB, \emph{i.e.}, a Feed Forward Network (FFN as shown in Fig.~\ref{fig:pipeline} (d) ), a Spectral-wise Multi-head Self-Attention (S-MSA), and two layer normalization. Details of S-MSA are given in Fig.~\ref{fig:pipeline} (e).

\subsection{Spectral-wise Multi-head Self-Attention}
Suppose $\mathbf{X}_{in} \in \mathbb{R}^{H\times W \times C}$ as the input of S-MSA, which is reshaped into tokens $\mathbf{X} \in \mathbb{R}^{HW \times C}$. Then $\mathbf{X}$ is linearly projected into \emph{query} $\mathbf{Q} \in \mathbb{R}^{HW \times C}$, \emph{key} $\mathbf{K} \in \mathbb{R}^{HW \times C}$, and \emph{value} $\mathbf{V} \in \mathbb{R}^{HW \times C}$: 
\vspace{0.2mm}
\begin{equation}
\mathbf{Q} = \mathbf{X}\mathbf{W}^\mathbf{Q}, \mathbf{K} = \mathbf{X}\mathbf{W}^\mathbf{K}, \mathbf{V} = \mathbf{X}\mathbf{W}^\mathbf{V},
\label{linear_proj}
\vspace{0.2mm}
\end{equation}
where $\mathbf{W}^\mathbf{Q}$, $\mathbf{W}^\mathbf{K}$, and $\mathbf{W}^\mathbf{V} \in \mathbb{R}^{C \times C}$ are learnable parameters; $biases$ are omitted for simplification. Subsequently, we respectively split $\mathbf{Q}$, $\mathbf{K}$, and $\mathbf{V}$ into $N$ \emph{heads} along the spectral channel dimension: $\mathbf{Q} = [\mathbf{Q}_1,\ldots,\mathbf{Q}_N]$, $\mathbf{K} = [\mathbf{K}_1,\ldots,\mathbf{K}_N]$, and $\mathbf{V} = [\mathbf{V}_1,\ldots,\mathbf{V}_N]$. The dimension of each head is $d_h = \frac{C}{N}$. Please note that Fig.~\ref{fig:pipeline} (e) depicts the situation with $N$ = 1 and some details are omitted for simplification. Different from original MSAs, our S-MSA treats each spectral representation as a token and calculates self-attention for $head_j$:
\vspace{-0.2mm}
\begin{equation}
\mathbf{A}_j = \text{softmax}(\sigma_j \mathbf{K}_j^\text{T} \mathbf{Q}_j), ~~{head}_j  =\mathbf{V}_j \mathbf{A}_j, 
\label{s-attention}
\vspace{-0.2mm}
\end{equation}
where $\mathbf{K}_j^\text{T}$ denotes the transposed matrix of $\mathbf{K}_j$. Because the spectral density varies significantly with respect to the wavelengths, we use a learnable parameter $\sigma_j \in \mathbb{R}^{1}$ to adapt the self-attention $\mathbf{A}_j$ by re-weighting the matrix multiplication $\mathbf{K}_j^\text{T} \mathbf{Q}_j$ inside $head_j$. Subsequently, the outputs of $N$ \emph{heads} are concatenated to undergo a linear projection and then is added with a position embedding:
\begin{equation}
\text{S-MSA}(\mathbf{X}) =\big(\mathop{\text{Concat}}\limits_{j=1}^{N}(head_{j})\big)\mathbf{W} + f_p(\mathbf{V}),
\label{agg_heads}
\end{equation}
where $\mathbf{W} \in \mathbb{R}^{C \times C}$ are learnable parameters, $f_p(\cdot)$ is the function to generate position embedding. It consists of two depth-wise \emph{conv}3$\times$3 layers, a GELU activation, and reshape operations. The HSIs are sorted by the wavelength along the spectral dimension. Therefore, we exploit this embedding to encode the position information of different spectral channels. Finally, we reshape the result of Eq.~\eqref{agg_heads} to obtain the output feature maps $\mathbf{X}_{out} \in \mathbb{R}^{H\times W \times C}$.

\begin{figure*}[t]
	\begin{center}
		\begin{tabular}[t]{c} \hspace{-2mm} 
			\includegraphics[width=0.97\textwidth]{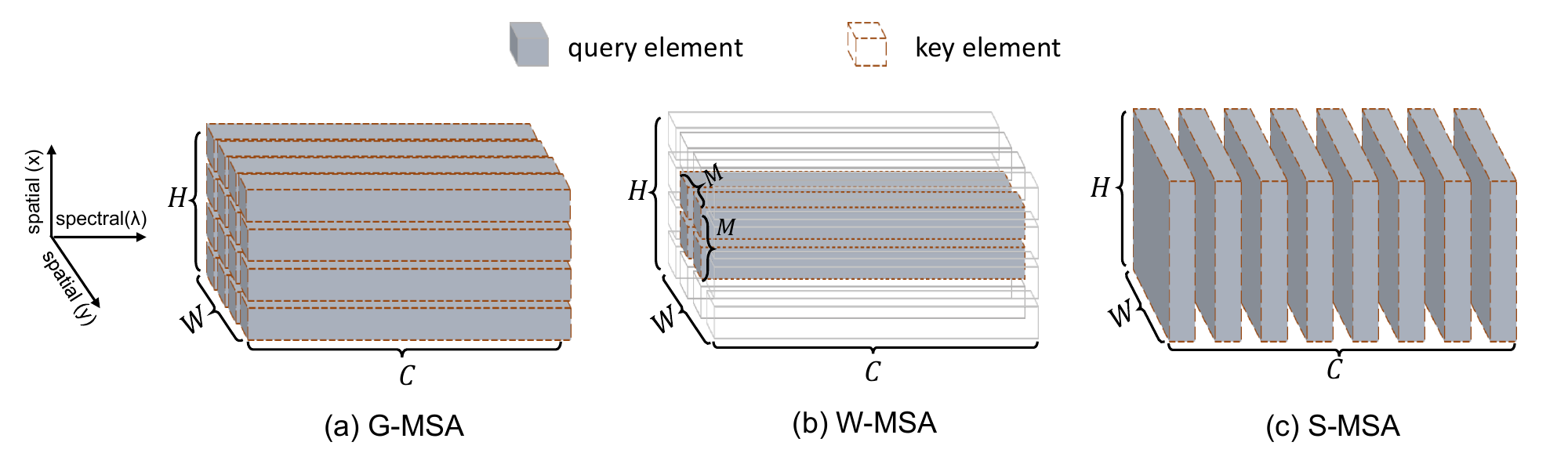}
		\end{tabular}
	\end{center}
	\vspace{-5mm}
	\caption{\small Diagram of different MSAs. The dark colored box represents $query$ element and the dashed box denotes $key$ element. (a) Global MSA samples all the tokens (pixel vectors) as $query$ and $key$ elements. (b) W-MSA calculates the self-attention inside position-specific windows. (c) The adopted S-MSA treats each spectral channel as a token and calculates the self-attention along the spectral dimension. }
	\label{fig:attan_compare}
	\vspace{-1mm}
\end{figure*}

\subsection{Discussion with Original Transformers}
\label{sec:discuss}
In this section, we  introduce the general paradigm of MSA in Transformer and then we analyze the computational complexity of the spatial-wise MSAs in original Transformers and the adopted S-MSA.

\subsubsection{General Paradigm of MSA}
We denote the input token as $\mathbf{X} \in \mathbb{R}^{n\times C}$, where $n$ is to be determined. In spatial-wise MSAs, $n$ denotes the number of tokens. In S-MSA, $n$ represents the dimension of the token. $\mathbf{X}$ is firstly linearly projected into \emph{query} $\mathbf{Q} \in \mathbb{R}^{n\times C}$, \emph{key} $\mathbf{K} \in \mathbb{R}^{n\times C}$, and \emph{value} $\mathbf{V} \in \mathbb{R}^{n\times C}$:
\begin{equation}
\mathbf{Q} = \mathbf{X} \mathbf{W^Q}, \mathbf{K} = \mathbf{X} \mathbf{W^K}, \mathbf{V} = \mathbf{X} \mathbf{W^V},
\label{eq:linear_proj}
\end{equation}
where $\mathbf{W^Q},\mathbf{W^K}$, and $\mathbf{W^V} \in \mathbb{R}^{C\times C}$ are learnable parameters; biases are omitted for simplification. Subsequently, we respectively split $\mathbf{Q}$, $\mathbf{K}$, and $\mathbf{V}$ into $N$ \emph{heads} along the spectral channel dimension: $\mathbf{Q} = [\mathbf{Q}_1,\ldots,\mathbf{Q}_N]$, $\mathbf{K}=[\mathbf{K}_1,\ldots,\mathbf{K}_N]$, and $\mathbf{V}=[\mathbf{V}_1,\ldots,\mathbf{V}_N]$. The dimension of each head is $d_h=\frac{C}{N}$. Then MSA calculates the self-attention for each $head_j$:
\begin{equation}
head_j = \text{MSA}(\mathbf{Q}_j,\mathbf{K}_j,\mathbf{V}_j).
\label{eq:msa}
\end{equation}
Subsequently, the outputs of $N$ $\emph{heads}$ are concatenated along the spectral dimension and undergo a linear projection to generate the output feature map $\textbf{X}_{out} \in \mathbb{R}^{n\times C}$: 
\begin{equation}
\textbf{X}_{out} = \big(\mathop{\text{Concat}}\limits_{j=1}^{N}(head_{j})\big)\mathbf{W},
\label{eq:msa_out}
\end{equation}
where $\mathbf{W} \in \mathbb{R}^{C\times C}$ are learnable parameters. Please note that some other contents such as the position embedding are omitted for simplification. Because we only compare the main difference between original spatial-wise MSAs and S-MSA, \emph{i.e.}, the specific formulation of Eq.~\eqref{eq:msa}. 

\begin{table}[t]
	\begin{center}
		
		\setlength{\tabcolsep}{2.5pt}
		\scalebox{0.90}{ \hspace{-1.5mm}
			\begin{tabular}{l c c c}
				\toprule
				\rowcolor{color3} MSA Scheme &~ Global MSA ~&~ Local W-MSA ~&~\bf S-MSA  ~\\
				\midrule
				Receptive Field &Global &Local &Global \\
				Complexity to $HW$ &Quadratic &Linear &Linear \\
				Calculating Wise &Spatial &Spatial &Spectral \\
				\bottomrule
		\end{tabular}}
		\caption{\small Comparisons of the properties of  different MSAs. }
		\label{tab:msa_compare}
	\end{center}\vspace{-5mm}
\end{table}

\subsubsection{Spatial-wise MSA}
The spatial-wise MSA treats a pixel vector along the spectral dimension as a token and then calculates the self-attention for each $head_j$. Thus, Eq.~\eqref{eq:msa} can be specified as
\begin{equation}
head_j = \mathbf{A}_j \mathbf{V}_j, ~~\mathbf{A}_j = \text{softmax}(\frac{\mathbf{Q}_j\mathbf{K}_j^T}{\sqrt{d_h}}).
\label{eq:spatial_msa}
\end{equation}
Eq.~\eqref{eq:spatial_msa} neads to be calculated for $N$ times. Therefore, the computational complexity of spatial-wise MSA is
\begin{equation}
O(\text{Spatial-MSA}) = N (n^2 d_h + n^2 d_h) = 2n^2 C.
\label{eq:cost_spatial_msa}
\end{equation}
The spatial-wise MSA is mainly divided into two categories: global MSA~\cite{global_msa} and local window-based MSA~\cite{liu2021swin}. Now we analyze these two kinds of MSAs.

\noindent\textbf{Global MSA.} As shown in Fig.~\ref{fig:attan_compare} (a), global MSA samples all the tokens as $key$ and $query$ elements, and then calculates the self-attention. Thus, the number of tokens $n$ ($key$ or $query$ elements) is equal to $HW$. Then, according to Eq.~\eqref{eq:cost_spatial_msa}, the computational complexity of global MSA is
\begin{equation}
O(\text{Global MSA}) = 2(HW)^2 C,
\label{eq:global_msa}
\end{equation}
which is quadratic to the spatial size of the input feature map. Global MSA enjoys a very large receptive field but its computational cost is nontrivial and sometimes unaffordable. Meanwhile, sampling redundant $key$ elements may easily lead to over-smooth results~\cite{xiangtl_gald} and even non-convergence issue~\cite{de_detr}. To cut down the computational cost, researchers propose local window-based MSA.

\noindent\textbf{Window-based MSA.} As depicted in Fig.~\ref{fig:attan_compare} (b), W-MSA firstly splits the feature map into non-overlapping windows at size of $M^2$ and samples all the tokens inside each window to calculate self-attention. Hence, the number of tokens $n$ is equal to $M^2$ and W-MSA is conducted $\frac{HW}{M^2}$ times for all windows. Thus, the computational complexity is 
\begin{equation}
O(\text{W-MSA}) =  \frac{HW}{M^2}(2(M^2)^2C) = 2M^2HWC, 
\label{eq:cost_w_msa}
\end{equation}
which is linear to the spatial size ($HW$). W-MSA enjoys low computational cost but suffers from limited receptive fields inside position-specific windows. As a result, some highly related non-local tokens may be neglected. 

Original spatial-wise MSAs aim to capture the long-range dependencies of spatial regions. However, the HSI representations are spatially sparse while spectrally similar and correlated. Capturing spatial-wise interactions may be less cost-effective than modeling the spectral-wise correlations. Based on this HSI characteristic, we adopt S-MSA.

\subsubsection{S-MSA}
As shown in Fig.~\ref{fig:attan_compare} (b), S-MSA treats each spectral feature map as a token and calculates the self-attention along the spectral dimension. Then Eq.~\eqref{eq:msa} is specified as 
\begin{equation}
\mathbf{A}_j = \text{softmax}(\sigma_j \mathbf{K}_j^\text{T} \mathbf{Q}_j), ~~{head}_j  =\mathbf{V}_j \mathbf{A}_j, 
\label{eq:s_msa}
\end{equation}
where $\mathbf{K}_j^\text{T}$ denotes the transposed matrix of $\mathbf{K}_j$. We note that the spectral density varies significantly with respect to the wavelengths. Therefore, we exploit a learnable parameter $\sigma_j \in \mathbb{R}^{1}$ to adapt the self-attention $\mathbf{A}_j$ by re-weighting the matrix multiplication $\mathbf{K}_j^\text{T} \mathbf{Q}_j$ inside $head_j$. Because S-MSA treats a whole feature map as a token, the dimension of each token $n$ is equal to $HW$. Eq.~\eqref{eq:s_msa} needs to be calculated $N$ times. Thus, the complexity of  S-MSA is 
\begin{equation}
O(\text{S-MSA}) = N (d_h^2 n + d_h^2 n) = \frac{2HWC^2}{N}.
\label{eq:cost_Spectra_msa}
\end{equation}
The computational complexity of W-MSA and S-MSA are linear to the spatial size ($HW$), which is much cheaper than that of global MSA (quadratic to $HW$). Nonetheless, S-MSA treats each spectral feature as a token. When calculating the self-attention $\mathbf{A}_j$, S-MSA views the global representations and $\mathbf{A}_j$ functions as global spatial positions. Therefore, the receptive fields of S-MSA are global and not limited to the position-specific windows. 

In addition, S-MSA calculates self-attention along the spectral dimension, which is based on HSI characteristics and more suitable for HSI reconstruction when compared to spatial-wise MSAs. Thus, S-MSA is considered to be more cost-effective than global MSA and W-MSA.

For brevity, we summarize the properties of global MSA, window-based MSA, and  S-MSA in Tab.~\ref{tab:msa_compare}. S-MSA enjoys global receptive fields, models the spectral-wise self-similarity, and requires linear computational costs.

\subsection{Ensemble Strategy}
\label{sec:ensemble}
In NTIRE 2022 Spectral Reconstruction Challenge, we adopt three ensemble strategies including self-ensemble, multi-scale ensemble, and Top-K multi-model ensemble to improve the performance and generality of our MST++. Now in this part, we describe them in details.

\subsubsection{Self-Ensemble}
The RGB input is flipped up/down/left/right or rotated 90°/180°/270° to be fed into the network. Subsequently, the outputs are transformed to the original state to be averaged. 

\subsubsection{Multi-scale Ensemble}
We respectively train our models with patches at size of 256$\times$256, 128$\times$128, and 64$\times$64. Then the outputs (whole images) are averaged to improve the restoration quality.

\subsubsection{Top-K Multi-model Ensemble}
We also train MIRNet~\cite{mirnet}, MPRNet~\cite{mprnet}, Restormer~\cite{restormer}, HINet~\cite{hinet}, and MST~\cite{mst} families. The Top-K performers are selected for SR. Then we conduct our Top-K multi-model ensemble to fuse these reconstructed HSIs as
\begin{equation}
\mathbf{Y}_{ens} = \sum_{i=1}^{\text{K}} \alpha_i \mathbf{\hat{Y}}^t_i,
\end{equation}
where $\mathbf{Y}_{ens} \in \mathbb{R}^{H\times W\times N_{\lambda}}$ denotes the ensembled HSIs, $\mathbf{\hat{Y}}^t_i$ represents the reconstructed HSIs of the $i$-th model, and $\alpha_i$ represents hyperparameter satisfying $\sum_{i=1}^{\text{K}} \alpha_i = 1$.

\section{Experiment}
\subsection{Dataset}
The dataset provided by NTIRE 2022 Spectral Reconstruction Challenge contains 1000 RGB-HSI pairs. This dataset is split into \texttt{train},  \texttt{valid}, and \texttt{test} subsets  in proportional to 18:1:1. Each HSI at size of 482$\times$512 has 31 wavelengths from 400 nm to 700 nm. To generate the corresponding RGB counterpart $\mathbf{I} \in \mathbb{R}^{H\times W\times 3}$, a transformation matrix $\mathbf{M} \in \mathbb{R}^{N_{\lambda}\times3}$ is applied to the ground-truth HSI cube $\mathbf{Y} \in \mathbb{R}^{H\times W\times N_{\lambda}}$ as 
\begin{equation}
	\mathbf{I} = \mathbf{Y} \times \mathbf{M}.
\end{equation}
Then the generated RGB images are injected with shot noise to simulate the real-camera situation. 

\subsection{Implementation Details}
During the training procedure, RGB images are linearly rescaled to [0, 1], after which $128\times 128$ RGB and HSI sample pairs are cropped from the dataset. The batch size is set to 20 and the parameter optimization algorithm chooses Adam modification with $\beta_1=0.9$ and $\beta_2=0.999$. The learning rate is initialized as 0.0004 and the Cosine Annealing scheme is adopted for 300 epochs. The training data is augmented with random rotation and flipping. The proposed MST++ has been implemented on the Pytorch framework and approximately 48 hours are required for training a network on a single RTX 3090 GPU. MRAE loss function between the predicted and ground-truth HSI is adopted as the objective. In the implementation of our MST++, we set $N_s$ = 3, $N_1$ = $N_2$ = $N_3$ = 1, $C$ = 31. 

During the testing phase, the RGB image is also linearly rescaled to [0, 1] and fed into the network to fulfill the spectral recovery. Our MST++ takes 102.48 ms for per image (size 482$\times$512$\times$3) reconstruction on an RTX 3090 GPU.

We adopt three evaluation metrics to assess the model performance. The first metric is mean relative absolute error (MRAE) that computes the pixel-wise disparity between all wavelengths of the reconstructed and ground-truth HSIs. MRAE can be formulated as
\begin{equation}
	\text{MRAE}(\mathbf{Y},\mathbf{\hat{Y}}) = \frac{1}{N} \sum_{i=1}^{N} \frac{\big|~\mathbf{Y}[i] - \mathbf{\hat{Y}}[i]~\big|}{\mathbf{Y}[i]},
\end{equation}
where $\mathbf{\hat{Y}} \in \mathbb{R}^{H\times W\times N_{\lambda}}$ indicates the reconstructed HSI cube and $N = H\times W\times N_{\lambda}$ denotes the number of all values on the image. The second metric is the root mean square error (RMSE) that is defined as
\begin{equation}
\text{RMSE}(\mathbf{Y},\mathbf{\hat{Y}}) = \sqrt{\frac{1}{N} \sum_{i=1}^{N} \big(\mathbf{Y}[i] - \mathbf{\hat{Y}}[i]\big)^2}.
\end{equation}
Since the deciding metric for the NTIRE 2022 Spectral Reconstruction Challenge is MRAE, we directly set it as the training objective for our SR  models. The last metric is the Peak Signal-to-Noise Ratio (PSNR).

\begin{table*}
	\begin{center}
		\setlength{\tabcolsep}{5.9pt}
		\resizebox{0.97\textwidth}{26mm}{\noindent
			\begin{tabular}{l | c c c c c | l | c c}
				\toprule[0.15em]
				 \multicolumn{6}{c|}{NTIRE 2022 HSI Dataset - \texttt{Valid}} &\multicolumn{3}{c}{NTIRE 2022 HSI Dataset - \texttt{Test}}\\
				 \midrule
				Method~~~~~~~~ & ~~Params (M)~~ & ~~FLOPS (G)~~ &~~~~~MRAE~~~~~ & ~~~~~RMSE~~~~~ &~~~~~~PSNR~~~~~~ &Username~~~~ &~~~~MRAE~~~~ &~~~~RMSE~~~~ \\
				\midrule[0.15em]
				HSCNN+~\cite{shi2018hscnn} & 4.65  & 304.45   & 0.3814   & 0.0588  &26.36   &pipixia &0.2434  &0.0411       \\
				HRNet~\cite{orange_cat} & 31.70  & 163.81   & 0.3476   & 0.0550 &26.89   &uslab &0.2377  &0.0391       \\
				EDSR~\cite{edsr} & 2.42  & 158.32   & 0.3277   & 0.0437 &28.29   &orange\_dog &0.2377  &0.0376       \\
				AWAN~\cite{awan} &4.04 &270.61 &0.2500 &0.0367 &31.22 &askldklasfj &0.2345 &0.0361  \\
				HDNet~\cite{hdnet} &2.66 &173.81  &0.2048 &0.0317 &32.13 &HSHAJii &0.2308 &0.0364    \\
				HINet~\cite{hinet} &5.21 &31.04  &0.2032 &0.0303 &32.51 &ptdoge\_hot &0.2107 &0.0365  \\
				MIRNet~\cite{mirnet} &3.75 &42.95 &0.1890 & 0.0274  &33.29 &test\_pseudo &0.2036  &0.0324   \\
				Restormer~\cite{restormer}  &15.11  &93.77 &0.1833  &0.0274  &33.40 &gkdgkd & 0.1935  &0.0322 \\
				MPRNet~\cite{mprnet}  &3.62 &101.59 &0.1817  &0.0270 &33.50 &deeppf & 0.1767 &0.0322 \\
				MST-L~\cite{mst}  &2.45  &32.07 & 0.1772  &0.0256 &33.90 &mialgo\_ls & 0.1247  &0.0257 \\
				\midrule
				\textbf{MST++}  & \textbf{1.62} & \textbf{23.05}  & \textbf{0.1645} & \textbf{0.0248} & \textbf{34.32} &\textbf{MST++*} & \textbf{0.1131} & \textbf{0.0231}  \\
				\bottomrule[0.15em]
		\end{tabular}}
		\caption{Comparisons with SOTA methods on NTIRE 2022 HSI datasets (\texttt{valid} and \texttt{test}). * represents  using ensembled models. }
		\label{tab:valid}
	\end{center}\vspace{-3mm}
\end{table*}

\begin{figure*}[t]
	\begin{center}
		\begin{tabular}[t]{c} \hspace{-4.8mm}
			\includegraphics[width=1\textwidth]{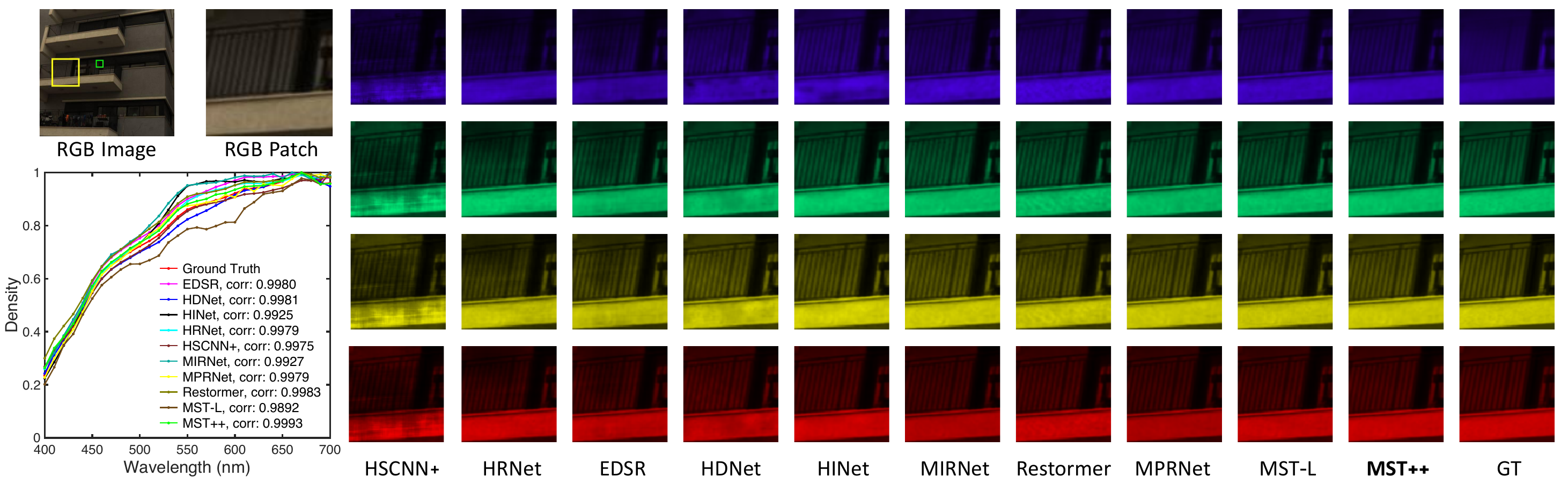}
		\end{tabular}
	\end{center}
	\vspace{-3mm}
	\caption{\small Reconstructed HSI comparisons of \emph{Scene} \texttt{ARAD\_1K\_0922} with 4 out of 31 spectral channels. 9 SOTA algorithms and our MST++ are included. The spectral curves (bottom-left) are corresponding to the selected green box of the RGB image. Please zoom in.}
	\label{fig:simulation}
\end{figure*}

\begin{table*}[t]
	\subfloat[\small Ablation study of different self-attention mechanisms.\label{tab:attention}]{\vspace{2mm}
		\scalebox{0.80}{
			\begin{tabular}{l c c c c c}
				\toprule[0.15em]
				Method &~~Baseline~~ &~~SW-MSA~~ &~~W-MSA~~ &~~G-MSA~~ &\bf~~S-MSA~~\\
				\midrule
				MRAE &0.3177 &0.2839 &0.2624 &0.1821 &\bf 0.1645  \\
				RMSE &0.0453 &0.0399 &0.0375 &0.0271 &\bf 0.0248 \\
				Params (M) &1.30 &1.60 &1.60 &1.60  &1.62  \\
				FLOPS (G)  &17.68 &24.10 &24.10 &25.11 &23.05 \\
				\bottomrule[0.15em]
	\end{tabular}}}\hspace{2mm}
	\subfloat[\small Ablation study of stage number $N_s$. \label{tab:stage}]{\vspace{2mm} 
		\scalebox{0.80}{
			\begin{tabular}{l c c  c c}
				\toprule[0.15em]
				$N_s$ &1 &2   &3 &4  \\
				\midrule
				 MRAE &~~0.1761~~ &~~0.1716~~ &\bf ~~0.1645~~ &~~0.1711~~  \\
				 RMSE &0.0266 &0.0269 &\bf 0.0248 &0.0265  \\
				 Params (M) &0.55 &1.08 &1.62 &2.16 \\
				 FLOPS (G) &8.10 &15.57 &23.05 &30.52 \\
				\bottomrule[0.15em]
	\end{tabular}}}
	\vspace{-1mm}
	\caption{\small Ablations. We train models on the \texttt{train} set and test on the \texttt{valid} set. MRAE, RMSE, Params, and FLOPS are reported.}
	\label{tab:ablations}\vspace{2mm}
\end{table*}

\begin{figure*}[t]
	\begin{center}
		\begin{tabular}[t]{c} \hspace{-2.7mm} 
			\includegraphics[width=1\textwidth]{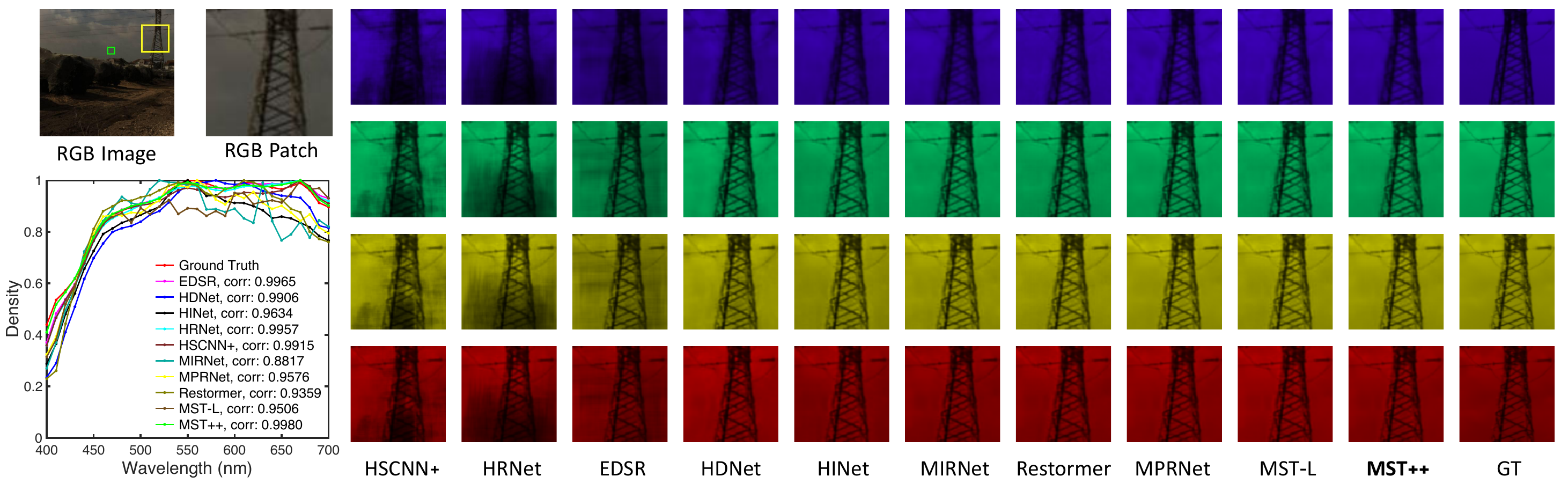}
		\end{tabular}
	\end{center}
	\vspace{-5mm}
	\caption{\small Reconstructed HSI comparisons of \emph{Scene} \texttt{ARAD\_1K\_0924} with 4 out of 31 spectral channels. 9 SOTA algorithms and our MST++ are included. The spectral curves (bottom-left) are corresponding to the selected green box of the RGB image. Please zoom in.}
	\label{fig:real}
\end{figure*}

\subsection{Main Results}
\subsubsection{Quantitative Results on \texttt{Valid} Set}
We compare our MST++ with SOTA methods including two SCI reconstruction methods (MST~\cite{mst} and HDNet~\cite{hdnet}), three SR algorithms (HSCNN+~\cite{shi2018hscnn}, AWAN~\cite{awan} and HRNet~\cite{orange_cat}), and five natural image restoration models (MIRNet~\cite{mirnet}, MPRNet~\cite{mprnet}, Restormer~\cite{restormer}, HINet~\cite{hinet}, EDSR~\cite{edsr}) on the \texttt{valid} set. Please note that HSCNN+~\cite{shi2018hscnn}, AWAN~\cite{awan} and HRNet~\cite{orange_cat} are the winners of NTIRE 2018~\cite{arad2018ntire} and 2020~\cite{arad2020ntire} Spectral Reconstruction Challenges. The results are listed in Tab.~\ref{tab:valid}. Our MST++ significantly outperforms SOTA methods by a large margin while requiring the least Params and FLOPS. For instance, our MST++ achieves 3.10, 7.43, and 7.96 dB improvement in PSNR while only requiring 40.10\% (1.62 / 4.04), 5.11\%, 34.84\% Params and 8.52\% (23.05 / 270.61), 14.07\%, 7.57\% FLOPS when compared to AWAN, HRNet, and HSCNN+.

To intuitively show the superiority of MST++, we provide PSNR-Params-FLOPS comparisons of different algorithms in Fig.~\ref{fig:teaser}. The vertical axis is PSNR (performance), the horizontal axis is FLOPS (computational cost), and the circle radius is Params (memory cost). It can be seen that our MST++ takes up the top-left corner, exhibiting the extreme efficiency advantages of our method.

\subsubsection{Quantitative Results on \texttt{Test} Set}
Tab.~\ref{tab:valid} lists the top-12 leaders of NTIRE 2022 Spectral Challenge (\texttt{test} set), where * indicates using ensembled models.  Impressively, our method won the championship out of 231 participants, suggesting the superiority of our MST++.

\subsubsection{Qualitative Results}
Fig.~\ref{fig:simulation} and~\ref{fig:real}  compares the reconstructed HSIs with 4 out of 31 spectral channels of nine SOTA methods and our MST++ on the \texttt{valid} set. Please zoom in for a better view. The top-left part depicts the input RGB image. The right part shows the reconstructed HSI patches of the selected yellow boxes in RGB image. It can be observed that previous methods show limitations in HSI detail restoration. They either achieve over-smooth HSIs  sacrificing fine-grained contents and structural details, or introduce unpleasing artifacts and blotchy textures. By contrast, MST++ does better in producing perceptually-pleasing and sharp-edge HSIs, and preserving  the spatial smoothness of the homogeneous regions. This is mainly because our MST++ excels at modeling inter-spectra self-similarity and  dependencies. Besides, the bottom-left part exhibits the spectral density curves  corresponding to the picked region of the green box in the RGB image. The highest correlation and coincidence between our curve and the ground truth verify the spectral-wise consistency restoration effectiveness of  MST++.

\subsection{Ablation Study}
we use the \texttt{valid} subset to conduct ablations. The baseline model is derived by removing S-MSA from  MST++.

\subsubsection{Self-Attention Mechanism}
We have discussed different self-attention mechanisms in Sec.~\ref{sec:discuss}. In this part,  we conduct ablation studies to verify the performance of these MSAs including global MSA (G-MSA)~\cite{global_msa}, local window-based MSA (W-MSA)~\cite{liu2021swin}, Swin MSA (SW-MSA)~\cite{liu2021swin}, and the adopted S-MSA~\cite{mst}. The results are reported in Tab.~\ref{tab:attention}. For fairness, the Params of models using different MSAs are set to the same value. Notely, the input feature of G-MSA is downscaled  into $\frac{1}{4}$ size to avoid out of memory.  It can be observed that our adopted S-MSA achieves the most significant improvement while requiring the least memory and computational costs. To be specific, when we respectively apply SW-MSA, W-MSA, G-MSA, and S-MSA, the performance is improved by 0.0338, 0.0553, 0.1356, and 0.1532 in MRAE while increasing 6.42, 6.42, 7.43, and 5.37 GFLOPS.  As analyzed in Sec.~\ref{sec:discuss}, these results mainly stem from the HSI spatially sparse while spectrally self-similar nature. Thus, capturing inter-spectra dependencies is more cost-effective than modeling correlations of spatial regions. 

\subsubsection{Stage Number}
We change the stage number $N_s$ of MST++ to investigate its effect. The results are shown in Tab.~\ref{tab:stage}. When $N_s$ = 3, the performance achieves its peak. Therefore, we finally adopt 3-stage MST++ as our SR model.

\subsubsection{Ensemble Strategy}
In Sec.~\ref{sec:ensemble}, we  adopt three ensemble strategies for NTIRE 2022 Spectral Reconstruction Challenge. In this part, we perform ablations to study their effects. On the \texttt{valid} set, self-ensemble, multi-scale ensemble, and Top-K (K is set to 5) multi-model ensemble respectively achieve improvements by 0.015, 0.033, and 0.045 in terms of MRAE. 

\section{Future Work}
Until now, there has not been a low-cost high-accuracy open-source baseline for SR research. Our MST++ aims to fill this gap. \textbf{Moreover, all the source code and pre-trained models in Tab.~\ref{tab:valid} (\texttt{valid}) including 11 SOTA methods are made publicly available.} Our goal is to provide a model zoo and toolbox to benefit the community.

\section{Conclusion}
In this paper, we propose the first Transformer-based framework, MST++, for spectral reconstruction from RGB. Based on the HSI spatially sparse while spectrally self-similar nature, we adopt S-MSA that treats each spectral feature map as a token for self-attention calculation to compose the basic unit SAB. Then SABs build up SST. Eventually, our MST++ is cascaded by several SSTs. Enjoying a multi-stage learning scheme, MST++ progressively improves the reconstruction quality from coarse to fine. Quantitative and qualitative experiments demonstrate that our MST++ dramatically  surpasses  SOTA methods while requiring cheaper memory and computational costs. Impressively, our MST++ won the \textbf{First} place in the NTIRE 2022 Challenge on Spectral Reconstruction from RGB.

\vspace{2mm}
\noindent \textbf{Acknowledgements:} This work is partially supported by the NSFC fund (61831014), the Shenzhen Science and Technology Project under Grant (ZDYBH201900000002, CJGJZD20200617102601004), the Westlake Foundation (2021B1501-2). Zudi Lin and Hanspeter Pfister acknowledge the support from NSF award IIS-2124179 and Google Cloud research credits.

{\small
\bibliographystyle{ieee_fullname}
\bibliography{reference}
}

\end{document}